\documentclass[journal]{IEEEtran}
\IEEEoverridecommandlockouts        

\usepackage[dvipsnames]{xcolor}	    
\usepackage[english]{babel}			
\usepackage[utf8]{inputenc}			
\usepackage{siunitx}                
\usepackage{epsfig}                 
\usepackage[hidelinks]{hyperref}    
\usepackage{graphics}               
\usepackage{float}	                
\usepackage{mathptmx}               
\usepackage{times}                  
\usepackage{amsmath}                
\usepackage{amssymb}                
\usepackage{cite}
\usepackage{esvect}                 
\usepackage{booktabs}				
\usepackage{tabularx}				

\usepackage[nameinlink,capitalise]{cleveref}    
\usepackage[all]{hypcap}                        

\usepackage{color, colortbl}
\usepackage{color}

\usepackage{newtxtext}
\usepackage{newtxmath}
\usepackage{calc}

\graphicspath{{Figures/}}
\DeclareGraphicsExtensions{.pdf,.jpg,.jpeg,.png}

\crefname{supfig}{Supplementary Fig.}{Supplementary Figs.}      
\Crefname{supfig}{Supplementary Figure}{Supplementary Figures}  

\newcommand{\finray}{Fin Ray\textsuperscript{\textregistered} }

\renewcommand{\vec}[1]{\vv{\boldsymbol{\mathrm{#1}}}}
\newcommand{\vecsub}[2]{\vv*{\boldsymbol{\mathrm{#1}}}{\,#2}}
\newcommand{\mat}[1]{\boldsymbol{\mathrm{#1}}}

\newcommand\copyrighttext{%
  \footnotesize \copyright~2023 IEEE. Personal use of this material is permitted. Permission from IEEE must be obtained for all other uses, in any current or future media, including reprinting/republishing this material for advertising or promotional purposes, creating new collective works, for resale or redistribution to servers or lists, or reuse of any copyrighted component of this work in other works.}
\newcommand\copyrightnotice{%
  \begin{minipage}{\textwidth}
  \copyrighttext
  \end{minipage}
}

\title{\LARGE \bf Avian-Inspired Claws Enable Robot Perching or Walking}

\author{Mohammad Askari$^{1}$\textsuperscript{*$\dagger$},
Won Dong Shin$^{1}$\textsuperscript{$\dagger$},
Damian Lenherr$^{1}$,
William Stewart$^{2}$\textsuperscript{$\dagger$},~\IEEEmembership{Member,~IEEE},
and Dario Floreano$^{1}$,~\IEEEmembership{Fellow,~IEEE}
\thanks{This work was supported by the Swiss National Science Foundation through the National Centre of Competence in Research (NCCR) and the European Union’s Horizon 2020 research and innovation program under grant agreement ID: 871479 AERIAL-CORE.}%
\thanks{$^{1}$Mohammad Askari, Won Dong Shin, Damian Lenherr, and Dario Floreano are with the Laboratory of Intelligent Systems, Swiss Federal Institute of Technology Lausanne (EPFL), CH-1015 Lausanne, Switzerland.}%
\thanks{$^{2}$William Stewart was with the Laboratory of Intelligent Systems, Swiss Federal Institute of Technology Lausanne (EPFL), CH-1015 Lausanne, Switzerland. He is now with the Soft Flyers Group, Stony Brook University, 11794, New York, USA.}%
\thanks{\textsuperscript{$\dagger$}These authors contributed equally to this work.}%
\thanks{\textbf{\textsuperscript{*}Corresponding author: mohammad.askari@epfl.ch}}%
}

\begin{document}

\maketitle

\makeatletter
\def\ps@IEEEtitlepagestyle{
  \def\@oddfoot{\copyrightnotice}
  \def\@evenfoot{}
}


\begin{abstract}

Multimodal UAVs (Unmanned Aerial Vehicles) are rarely capable of more than two modalities, i.e., flying and walking or flying and perching. However, being able to fly, perch, and walk could further improve their usefulness by expanding their operating envelope. For instance, an aerial robot could fly a long distance, perch in a high place to survey the surroundings, then walk to avoid obstacles that could potentially inhibit flight. Birds are capable of these three tasks, and so offer a practical example of how a robot might be developed to do the same. In this paper, we present a specialized avian-inspired claw design to enable UAVs to perch passively or walk. The key innovation is the combination of a Hoberman linkage leg with \finray claw that uses the weight of the UAV to wrap the claw around a perch, or hyperextend it in the opposite direction to form a curved-up shape for stable terrestrial locomotion. Because the design uses the weight of the vehicle, the underactuated design is lightweight and low power. With the inclusion of talons, the 45\,g claws are capable of holding a 700\,g UAV to an almost 20-degree angle on a perch. In scenarios where cluttered environments impede flight and long mission times are required, such a combination of flying, perching, and walking is critical.

\end{abstract}



\section{Introduction}

\IEEEPARstart{R}{ecently}, there has been a lot of interest in perching UAVs (Unmanned Aerial Vehicles). The advantages of perching include being able to use less energy than when maintaining flight, having the opportunity to recharge or refuel, and enabling long-term surveillance. Studies on UAV-perching have produced a diverse set of perching strategies such as sitting on poles and branches \cite{passive_perch}, hanging from branches \cite{hanging_branches_conference}, passively activated mechanisms \cite{mclarenPassiveClosingTendon2019}, perching on walls \cite{SCAMP-Overview}, perching on ceilings \cite{ceiling_effects}, or even just resting \cite{perching_resting_paradigm}. Others have considered the mechanics of perching, such as through dry adhesion\cite{dry_adhesion}, electrostatic adhesion \cite{ electrostatic_adhesion}, glues \cite{AirForceStickyPad}, and cables \cite{spiderMAV}. Work on perching has covered different UAV classes such as multicopters \cite{passive_perch}, fixed-wings \cite{stewartPassivePerchingEnergy2021}, and flapping-wings \cite{zuffereyHowOrnithoptersCan2022}.

Robotic flying-walking hybrid locomotion has been shown to be feasible on flat obstacle-free ground \cite{duck}. DUCK consisted of a multicopter with legs that could walk on the ground, take off, land, and fly repeatedly, but could not perch. One recent demonstration by another robot, LEONARDO, was dexterous enough to \lq slackline\rq \cite{slackline_scirob}. This robot used simple feet combined with active control from the quadcopter motors for balance and disturbance rejection. As a result, it requires constant power to remain in one spot. For search and rescue missions, robots could be required to hold position for long periods of time, acting as communication relays or fixed sensors for hours, which would require heavy batteries. Having a way to combine aerial-legged multimodality with passive perching could enable robots to operate for extended periods compared to fully actuated robots of similar mass.

\begin{figure*}[t]
\centering
\begin{center}
\includegraphics[width=1.0\textwidth]{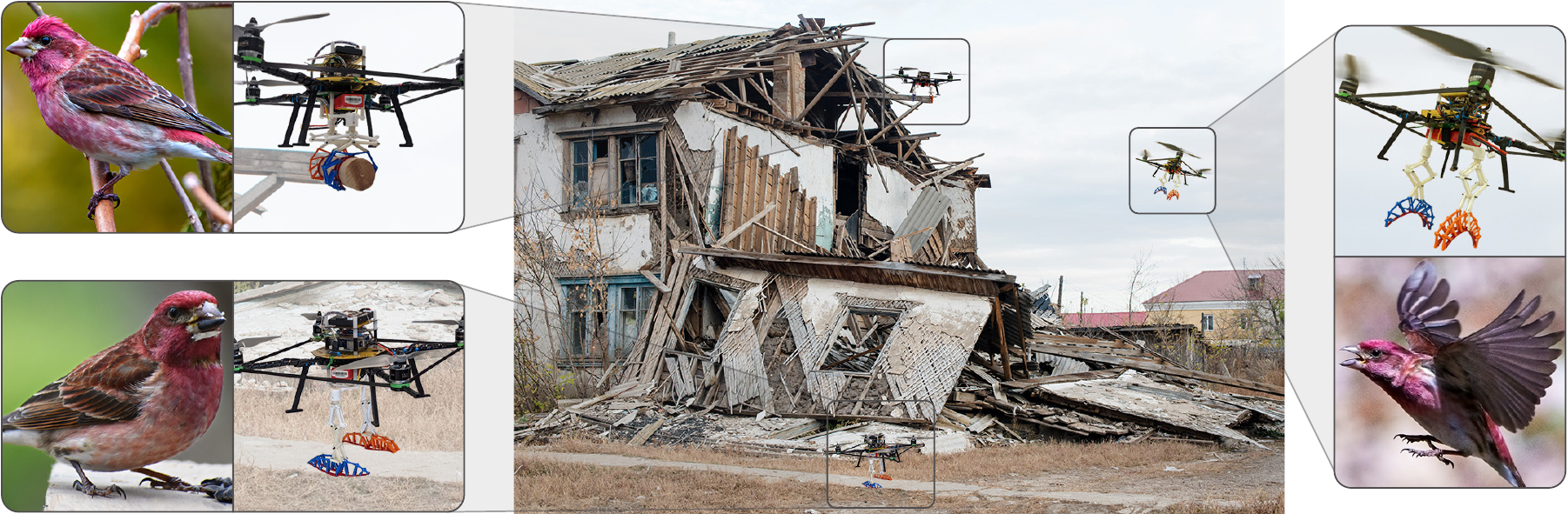}
\vspace{-0.55cm}
\caption{Sample mission of a flying-perching-walking robot for search and rescue. The inset views show comparisons of the feet of a purple finch and the robotic equivalent in different configurations. Photo credits: Rejean Aline; Claude Laprise; Olga Pink - Adobe Stock.}
\vspace{-0.65cm}
\label{figure1}
\end{center}
\end{figure*}

In working towards this goal, we focus on claw design to imbue a flying robot with the ability to perch passively or walk. In terms of perching, we focus on sitting atop horizontal poles, similar to \cite{passive_perch}. Inspired by the feet of birds, we propose a novel \finray \cite{pfaff2011application} claw design for a UAV that can perch passively on a horizontal pole or walk on the ground (Fig. \ref{figure1}). When the weight of the UAV presses on it, the claw passively wraps around and holds the perch. The weight is removed from the claw at takeoff, allowing it to slip off the perch. The claw can be used in a hyperextended state, in which the weight of the UAV instead causes the claw to curl upward. This can then enable the vehicle to walk around. We characterize the limits of the passive perching, showing that a \SI{700}{\gram} UAV can lean at 19.4 degrees before it slips. This performance improves on previous leaning experiments by about 5 degrees while also reducing the weight by three quarters \cite{passive_perch}. When in hyperextended mode, the upward-curled claw with low curvature provides a large support polygon, enabling a stable gait that is not possible with a claw that can only curl around an object to perch. We also measure the claw's squeezing force under similar conditions and find that it can squeeze with a force up to its body weight. For comparison, birds can squeeze with a force of up to two times their body weight but rely on muscles to achieve that performance \cite{bird_perch_reliability}.

\section{Related Work}

There are several examples of aerial-ground hybrid locomotion. These include a squirrel-inspired robot that can glide in the air, land, and then crawl under obstacles \cite{wondong_bioinspiration}; a winged UAV with rotating winglets that can push itself along the ground after landing \cite{daler_flying_walking}; A winged robot that uses mini-whegs to run on the ground \cite{boriaSensorPlatformCapable2005}. Others implemented a round cage to allow them to roll around the ground \cite{rolling_cage}. For simplicity, some aerial-ground vehicles just use wheels \cite{actuated_wheels_quad}. However, due to the nature of these modes of ground locomotion, the above examples often have a hard time overcoming obstacles. 

The most widely used foot designs in legged robots are ball-shaped feet and flat feet \cite{kaslin2018towards}. Ball-shaped feet do not limit the orientation of the feet with respect to the ground \cite{kaslin2018towards}, but they suffer from a limited contact area. This makes them ineffective at exerting a moment on the ground. On the other hand, flat feet provide a wide contact area but are limited in foot orientation. Ball-shaped feet are generally used in robots with four or more legs where using a moment with the ground is not as important as freedom of orientation. Flat feet are used in robots with one or two legs and are used to balance in moments when standing still. However, both these foot designs are only used for legged locomotion and not for grasping, perching, or other purposes.

If we turn to nature, we can see that birds are quite adept at perching, walking, and flying. Roderick et al. showed that when perching, Pacific Parrotlets use both friction between the pads of their feet in addition to talons to grip the perch \cite{bird_perch_reliability}. Furthermore, they found that depending on the perch material and diameter, the relative importance of the talons or pads of their feet changes. In some cases, the force due to the talons can reach up to eight times that of the pads of their feet. 

For many years, it was believed that birds utilize passive mechanisms to hold onto perches for long periods of time or while sleeping \cite{gill2007ornitology}. The basis of this hypothesis was a passive mechanism in the birds claw consisting of tendons that pass behind the ankle and are pulled when the birds squat. This would enable birds to remain perched when resting or even sleeping. However, Galton et al. cast doubt on this theory using experimental data showing that European Starlings could adapt to surgically severed tendons and sleep on perches \cite{galton2012experimental}. They also found that when anesthetized, the birds would fall from perches, which would not happen if there was a passive mechanism for perching. Nonetheless, many research groups have developed quadrotor perching mechanisms based on the incorrect tendon notion \cite{passive_perch,nadan2019bird}, indicating that the mechanism could be useful even if birds are not utilizing it. 

The mechanisms designed by these groups use stiff link segments (phalanges) connected by flexible joints \cite{passive_perch,nadan2019bird}. The tendon is an inelastic cable that transmits the weight of the vehicle into the phalanges, providing the actuation force. These underactuated designs allow the individual fingers to passively conform around objects of different shapes. The two designs in \cite{passive_perch,nadan2019bird} differ in the path of the tendon. As an alternative to using tendons to achieve passive perching, other authors \cite{chi2012finrayperching} used a slider mechanism in combination with \finray \cite{pfaff2011application} digits in an opposing arrangement \cite{chi2012finrayperching} that wrap around convex structure. Indeed, \finray fingers are also used in many compliant grippers \cite{shan2020finraymodelling,basson2019finray,elgeneidy2019finray,culler2012finrayperching,chi2012finrayperching,crooks2017passive}. Under load, the \finray finger structure bends in the opposite direction of the applied force and thus conform to the object. In the perching mechanism developed by Chi et al., digits were equipped with aluminum talons at the extremities of the \finray structure \cite{chi2012finrayperching}. The weight of the vehicle would close the \finray digits and press the talons into the perch. A similar active \finray claw has been developed for flapping wing MAVs, which weighs only 45g \cite{broersDesignTestingBioinspired2022}. This claw was shown to be effective on a variety of different-shaped perches, on slanted perches, and when approaching the perch from an angle. It is, however, quite limited in that any deviation from perfectly vertical perching after landing would cause the claw to fail.

Another approach to passive perching is the Sarrus mechanism \cite{burroughs2016sarrus}, where the weight of the UAV causes hinged plates to fold on themselves. These closing hinges are rigidly connected to plates such that when they close, the plates are brought together, squeezing the perch. At takeoff, the body weight is lifted off the mechanism, allowing the hinged plates to open and passively release the perch. While the plates are effective at transmitting the forces to the perch, the robot is likely to slip when leaning over. This perching mechanism does not use tendons to transfer forces.

To date, none of the claws developed for perching have been used also for walking. Indeed, many of these passive claws only actuate themselves to curl around an object, which would inhibit walking performance. The claw design described here leverages the weight of the vehicle for passive perching, but also enables inclined stable perch, and most importantly allows the flying vehicle to walk on the ground.


\section{Leg and Claw Design}

\begin{figure*}[t]
\centering
\begin{center}
\includegraphics[width=1.0\textwidth]{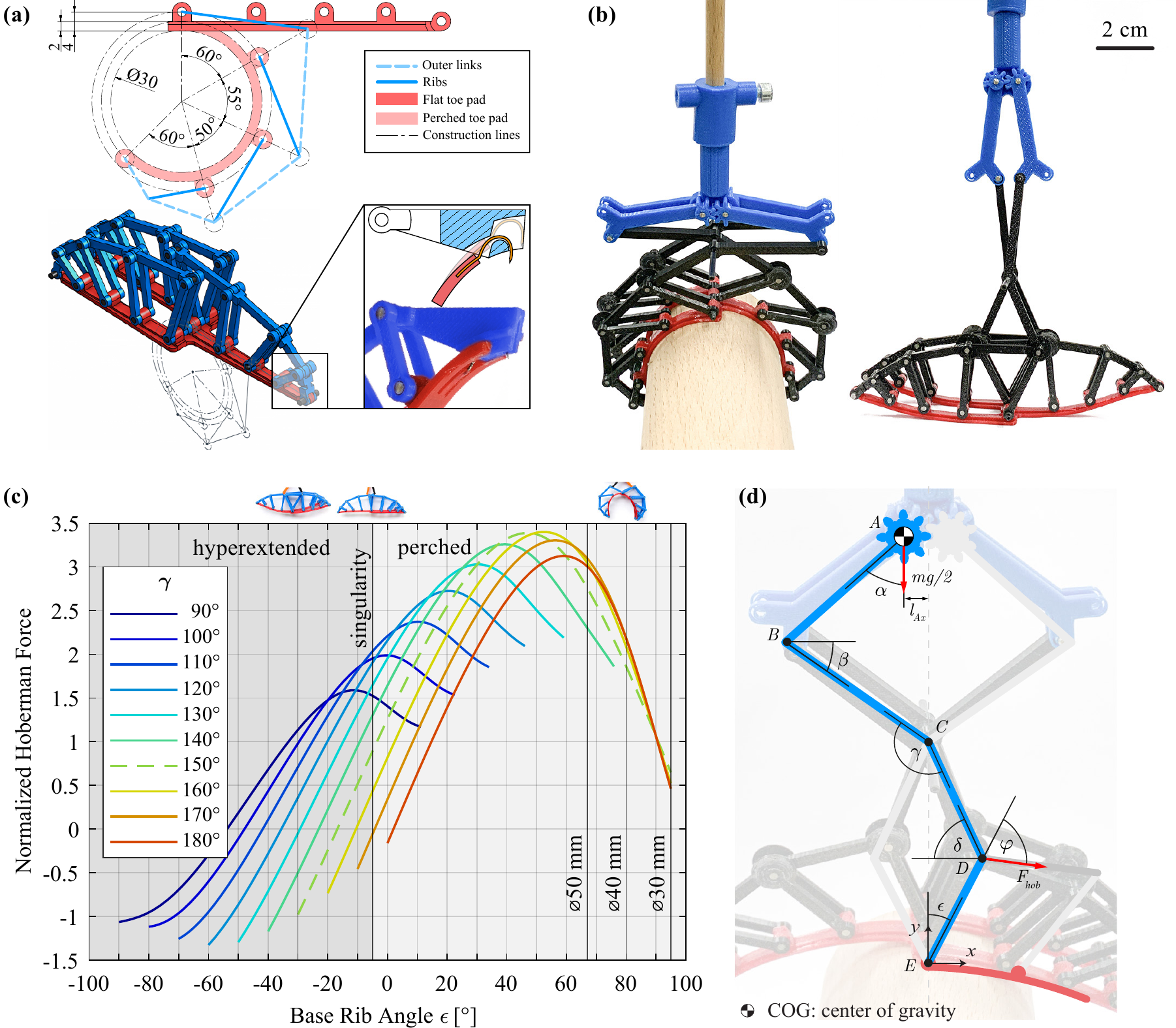}
\vspace{-0.5cm}
\caption{Leg and claw design. \textbf{(a)} Geometric parameter sizing of the claw is done in the perched configuration. The CAD of the claw was sized based on the calculations presented in the bottom of the figure. \textbf{(b)} Photos of the assembled claw in perched (left) and hyperextended (right) modes. \textbf{(c)} Mechanical advantage profiles of the Hoberman linkage over a range of base rib angles ($\epsilon$) for different $\gamma$ values. Grey shading shows whether the claw is within the perched or the hyperextended region. The vertical black line at -5 degrees indicates the singularity point where the claw switches between the two modes. The other vertical lines show the limits of claw geometry ($\gamma = \SI{150}{\degree}$) selected for this project. \textbf{(d)} Diagram and geometric parameters of the linkage.}
\vspace{-0.6cm}
\label{claw}
\end{center}
\end{figure*}

The mechanical design combines a Hoberman linkage \cite{HobermanRadialExpansionRetraction1991} leg with a \finray claw. The combined structure has two stable configurations, perched and hyperextended. In the perched configuration, the \finray claw is curled inward, and in the hyperextended configuration, the claw is stretched out. We manually switch between perched and hyperextended modes by positioning the Hoberman linkage leg in a collapsed (perched) or stretched form (hyperextended). When moving from one configuration to another, the \finray claw passes through a singular point resembling a nearly flat claw. The moment the claw passes this singularity, the weight of the UAV will passively push the claw to either configuration.

\begin{figure*}[t]
\centering
\begin{center}
\includegraphics[width=1.0\textwidth]{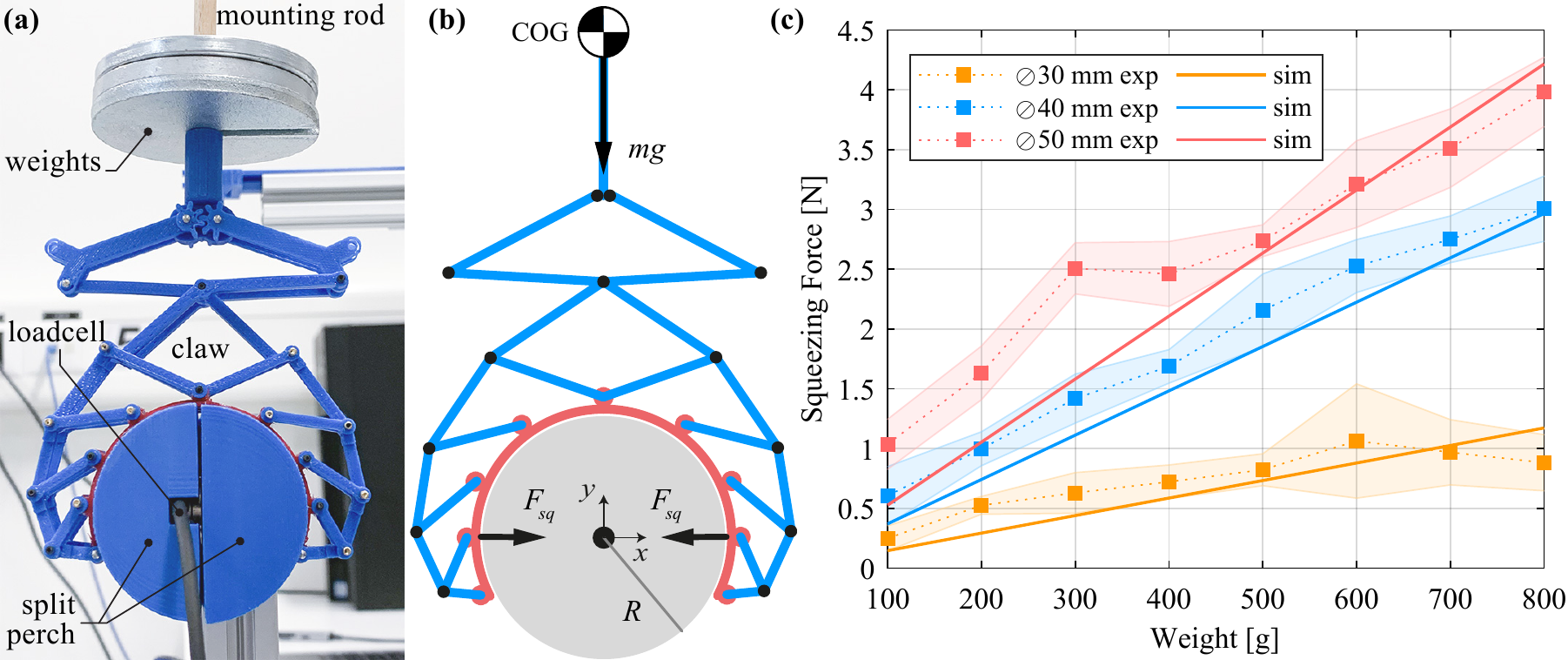}
\vspace{-0.525cm}
\caption{\textbf{(a)} Split perch experimental setup. Two halves of the split perch encompass the ATI Nano17 loadcell. Weights are mounted to a wooden rod, which press on the talon-less claw, which converts the weight into a squeezing force measured by the load cell. \textbf{(b)} Diagram of the claw in perched configuration. \textbf{(c)} Plot of the estimated and measured squeezing force as a function of weight. The shaded regions represent the standard deviation of 10 measurements.}
\vspace{-0.6cm}
\label{split_perch}
\end{center}
\end{figure*}

The claw's design is inspired by the anisodactyl toe arrangement found in perching birds, where three digits (toes) face forward and one backward. The forward middle toe is omitted, leaving space for the backward‐oriented toe, to overlap between the remaining front digits when wrapping around smaller perches. The design methodology involves selecting the claw linkage dimensions in the perched configuration based on the smallest desired perch size. Figure \ref{claw}a shows the sagittal view of the middle toe pad together with the dimensioned sketch on the desired minimum perch size of \SI{30}{\milli\meter} in diameter. The hubs of the toe pad are spaced \SI{4}{\milli\meter} above the surface (\SI{2}{\milli\meter} at the tip). In the perched configuration, they are arranged at angles of \SI{60}{\degree}, \SI{55}{\degree}, \SI{50}{\degree}, and \SI{60}{\degree} relative to the vertical axis starting from the first hub, which results in a \SI{90}{\degree} overlap. These angles determine the sizes of the claw phalanges from proximal to distal (see Supplementary Text for detailed information on linkage dimensions). In order to accommodate both walking and perching behaviors, we selected the angles to strike a balance between the equal phalanges found in perching birds and the long-to-short phalanges seen in walking birds as we move from proximal to distal \cite{kavanagh2013developmental}. The palm of the claw is made of a flexible, 3D-printed TPU (Thermoplastic Polyurethane) toe pad (red in Fig. \ref{claw}a). The ribs, 3D-printed in ABS (Acrylonitrile Butadiene Styrene), connect each toe pad hub to the outer-link of the \finray mechanism (blue structure in Fig. \ref{claw}a), positioned radially above the subsequent hub. To increase perching stability, the claw includes talons integrated at both extremities of the rib structure (inset of Fig. \ref{claw}a). When a horizontal force is exerted on the central rib of each digit, the \finray ribs and outer-links transfer the loads along the digit, causing the digit to curl. Depending on the direction of the force, the digits will either curl downward, closing around a perch, or stretch upward into the hyperextended mode creating a foot shape suitable for walking (left and right respectively in Fig. \ref{claw}b).

The geometry of the Hoberman linkage regulates the curling angle of the \finray claw. In particular, the angle of the linkage ($\gamma$ in Fig. \ref{claw}d) affects the magnitude of the Hoberman force and the resulting curling angle of the claw ($F_{hob}$ and $\epsilon$ in Fig. \ref{claw}d, respectively). The output force imparted on the \finray claw by the weight of the UAV can be modeled with:
\begin{equation}
F_{hob} = \frac{\sin\epsilon + \cos\epsilon\left(\frac{l_{BC}}{l_{CD}}\left(\frac{\cos\beta}{\sin\delta}+\tan\alpha\frac{\sin\beta}{\sin\delta}\right) + \frac{\cos\delta}{\sin\delta}\right)}{\sin\varphi}mg/2 \ ,
\label{hoberman_equation}
\end{equation}
\noindent where AB, BCD, and DE are the individual links and $\alpha$, $\beta$, $\gamma$, $\delta$, $\epsilon$, and $\phi$ are the corresponding angles (Fig. \ref{claw}d). The two most critical angles are $\gamma$ and $\epsilon$, and the rest are geometrically dependent on these two. This is because $\gamma$ is set when designing the claw and does not vary, and $\epsilon$ is a direct measure of how far the claw can curl or stretch. Using equation \ref{hoberman_equation}, we calculated $F_{hob}$ for a variety of $\gamma$ and $\epsilon$ values. $\gamma$ was varied between 90 and 180 degrees in increments of 10 degrees. A $\gamma$ value of 150 degrees was selected for the claws developed in this work. (Fig. \ref{claw}c). This is because, at 150 degrees, the claw balances an ability to reach the hyperextended state (dashed line at $\epsilon=-30$ degrees in Fig. \ref{claw}c) as well as being able to curl around a perch of \SI{30}{\milli\meter} diameter (dashed line at $\epsilon=-95$ degrees in Fig. \ref{claw}c). In addition, the maximum achievable Hoberman Force occurs at $\gamma=150$ degrees. Changing $\gamma$ shifts the achievable range of $\epsilon$ values. For example, increasing $\gamma$ would allow the claw to reach a higher maximum $\epsilon$, which corresponds to being able to perch on smaller diameter perches (This corresponds to the right end of the red $\gamma=180$ degrees line in Fig. \ref{claw}c being at $\epsilon=95$ degrees whereas the right end of the blue $\gamma=110$ degrees being at only 35 degrees). The drawback is that the higher $\gamma$ would limit the hyperextended stretching (notice that the red line in Fig. \ref{claw}c cannot even reach the hyperextended state), which is required for walking. On the other hand, decreasing $\gamma$ would allow for more hyperextended stretching, but would limit the range of perching diameters. 

A single \SI{10.2}{\cm} long claw and Hoberman leg weighs only \SI{23}{\gram}. When servos, leg linkages, and a controller are added to the leg to enable walking, the total weight goes up to \SI{57}{\gram}. The claws are used as a pair, so the whole system comes to \SI{114}{\gram}. This is considerably lighter than the designs by Doyle and Nadan (\SI{478}{\gram} and \SI{178}{\gram}, respectively) \cite{passive_perch,nadan2019bird}.


\section{Perching Characterization}

\begin{figure*}[htp]
\centering
\begin{center}
\includegraphics[width=1.0\textwidth]{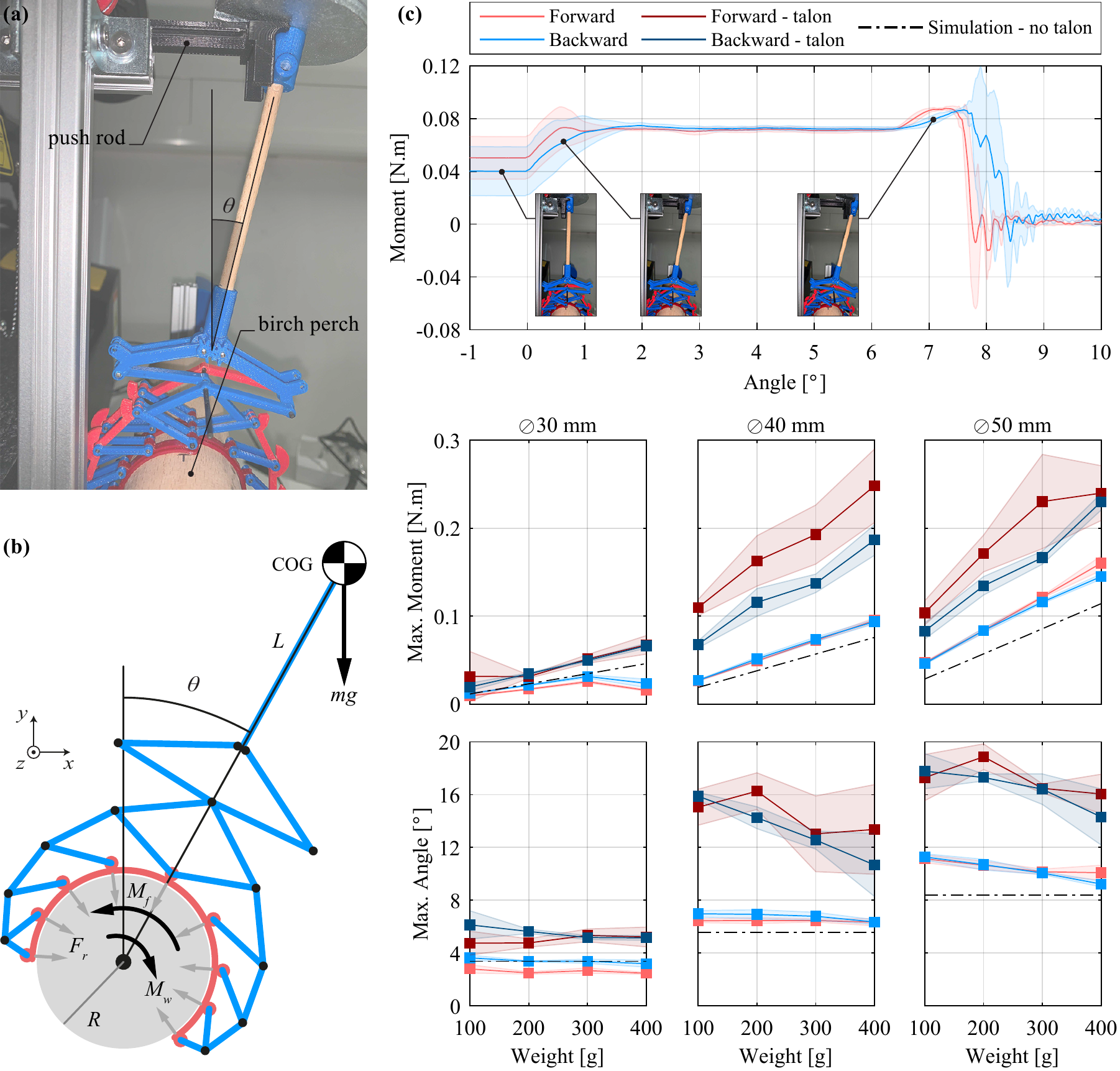}
\caption{\textbf{(a)} Slip resistance experimental setup. \textbf{(b)} Schematic force diagram of the model of the experiment (see Supplementary Text for more details). This illustration shows how the moment due to the mass of the UAV ($M_w$) is counteracted by the moment due to the friction of the claw ($M_f$) grasping the perch. \textbf{(c)} Maximum tilting angle and moment measurements for different weights and perch diameters. The plot at the top shows variation in the moment over time for the case of \SI{300}{\gram} weight on a \SI{40}{\milli\meter} perch. The shaded regions represent the standard deviation of repeated experiments.}
\label{tilting_characterization}
\end{center}
\end{figure*}

\begin{figure*}[htp]
\centering
\begin{center}
\includegraphics[width=1.0\textwidth]{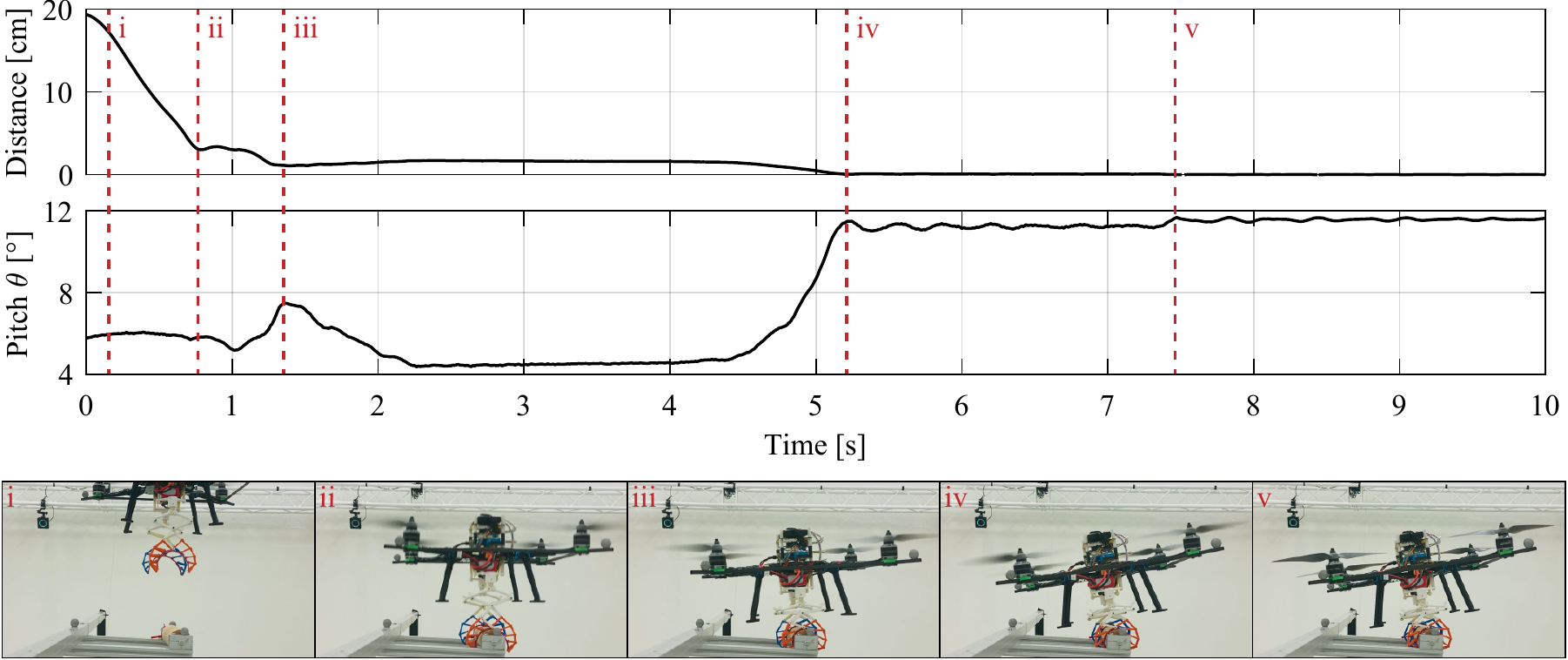}
\vspace{-0.5cm}
\caption{Perching experiment with a manually piloted quadcopter equipped with a set of two claws. (top) vertical distance from the perch and (middle) the pitch angle over time. (bottom) Snapshots from the video corresponding to different instances of the perching maneuver. Instant (i) shows the UAV in flight before touchdown, and (ii) it makes contact with the wooden perch. (iii) shows when the claws are fully closed and firmly grip the perch. At (iv) the UAV tilts back when the thrust is significantly reduced, and when the propellers are completely stopped (v), the UAV remains perched at 11.6 degrees.}
\vspace{-0.6cm}
\label{perch_test}
\end{center}
\end{figure*}

Variations in the perching approach and maneuver can lead to the UAV tilted to one side of the perch. To be robust while perching, the claw must maintain its grip even when the UAV is tilted. We characterized this robustness by measuring the squeezing force a single claw can exert, how far the UAV can tilt before falling, and how much torque the claw can exert before slipping. We also demonstrated the robust perching in flight with a UAV equipped with a set of two claws.

To measure the squeezing force of the claw, we used a 3D printed, ABS split perch setup similar to the one in \cite{bird_perch_reliability} (Fig. \ref{split_perch}a). We found that the squeezing force is directly proportional to body weight and perch diameter (Fig. \ref{split_perch}c). This matches the predictions of the analysis of the Hoberman linkage (Fig. \ref{claw}c), which indicate that the force imparted on the \finray device by the Hoberman linkage increases with increasing perch diameter. For comparison, this trend of lower squeezing forces on smaller diameter perches is the opposite of what was found by Roderick et al. \cite{bird_perch_reliability}, who noted that the birds squeezed harder on smaller perches. Furthermore, the birds could squeeze up to two times their body weight. The maximum normalized squeezing force generated by the passive claw was just over one, with \SI{100}{\gram} of weight on the \SI{50}{\milli\meter} perch. Reaching a squeezing force comparable to the birds would require either the addition of actuators or modifying the Hoberman link to provide more mechanical advantage.

The results of the squeezing force characterization are compared to a static force model, detailed in Supplementary Text. The assumptions made in modeling the claw prove reasonable, as the model predicts well the trends in the experimental results (Fig. \ref{split_perch}c). In addition, it correctly estimates that the increase in squeezing force from \SIrange{40}{50}{\milli\meter} is less than the increase in squeezing force from \SIrange{30}{40}{\milli\meter}. However, the prototype claw overperforms the model at lower weights, especially for the \SI{50}{\milli\meter} perch. Under these lower weights, the claw experiences less deformation in the toe pad and does not perfectly conform to the shape of the perch, which is explicitly not accounted for in the model.

\begin{figure*}[htbp]
\centering
\begin{center}
\includegraphics[width=1.0\textwidth]{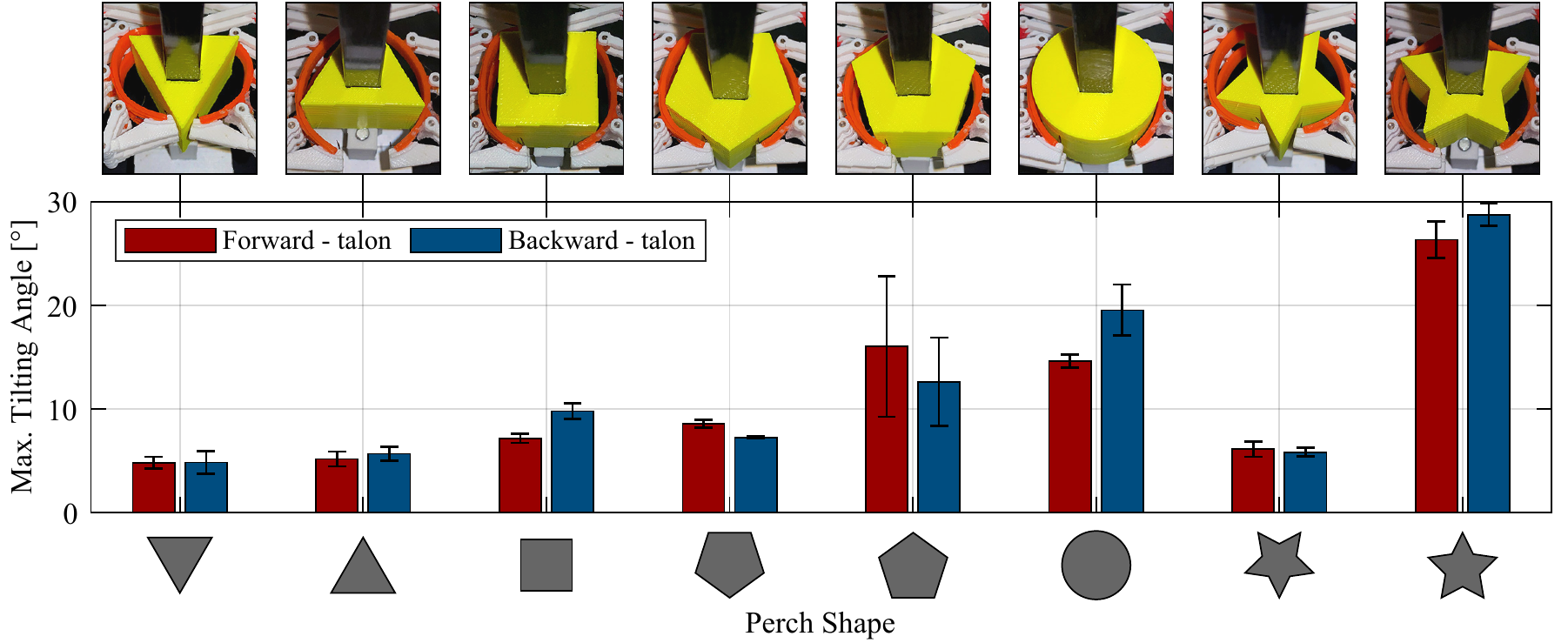}
\vspace{-0.5cm}
\caption{Maximum sustained angle of the quadcopter with two claws on different perch shapes. The error bars show the standard deviation of five tests.}
\vspace{-0.6cm}
\label{irregular perches}
\end{center}
\end{figure*}

We investigated how far off-center the claw can reach before slipping by setting up a benchtop test apparatus to measure the maximum sustainable angle and corresponding moment (Fig. \ref{tilting_characterization}a). The claw was vertically placed on a solid birch perch ($\theta=0$) with weights mounted at a distance of \SI{200}{\milli\meter} above the rotation axis of the perch. Using an Instron universal testing machine, a push rod pushes the arm and increases the tilting angle $\theta$. A 6-axis load cell (ATI Gamma), connected to one end of the perch, measures the moment created by the claw. Experiments are considered quasi-static due to the slow pushing rate of \SI{5}{\milli\meter\per\second}. Ten tests were done for each weight and perch diameter combination, with the orientation of the claw being switched after five tests to capture effects due to its asymmetric design. We collected data from the load cell at a frequency of \SI{1000}{\hertz} and used a moving average filter with a window size of $100$ samples to remove noise. Finally, the experiments were repeated with a talon-less claw to understand the effects of the talons. 

Representative moment data from one of the experiments is shown in Fig. \ref{tilting_characterization}c alongside predictions from the static force model, detailed in Supplementary Text. There are three main phases of the test. In the beginning, the wooden rod is vertical, and the push rod is not in contact with it. At $\theta=0$, the push rod comes into contact with the wooden rod, marking the beginning of the second phase. In this phase, the wooden rod is pushed, and the moment increases. During this phase, static friction holds the claw in position against the increasing moment. The measured increase in $\theta$ during this period results from the claw deforming. At about 2 degrees, the moment levels off and remains constant. During this phase, the claw is slipping, and dynamic friction is holding the claw on the perch. Once the claw reaches about 6.5 degrees, the moment due to the weight begins increasing again. Lastly, there is a sharp drop in the moment as the claw finally slips, and just before 8 degrees, the wooden rod falls and hits a catch.
 
Increasing the weight causes a greater measured moment before slipping, as expected based on the results of the squeezing test (Fig. \ref{tilting_characterization}c). However, the maximum angle is less affected by weight change and remains almost constant. The claw performs better on larger perch sizes due to increased squeezing force with perch diameter. The model predicts these trends well but with minor errors attributable to decreasing model accuracy at lower weights, as previously discussed. Unlike squeezing force experiments, the forces at the forward and backward digit joints differ in the tilting tests. At its current state, the claw can sustain weights up to \SI{400}{\gram} (twenty times its own weight). The 3D-printed joints and the claw would require reinforcement and design optimization to reach weights beyond this point. With the addition of talons to the claw, there is a considerable increase (50-150\%) in the achievable angle. However, it comes with an increase in variability between runs. This is due to the talons getting caught in asperities, which are randomly dispersed in the grain of the wooden perch.

\begin{figure*}[htbp]
\centering
\begin{center}
\includegraphics[width=1\textwidth]{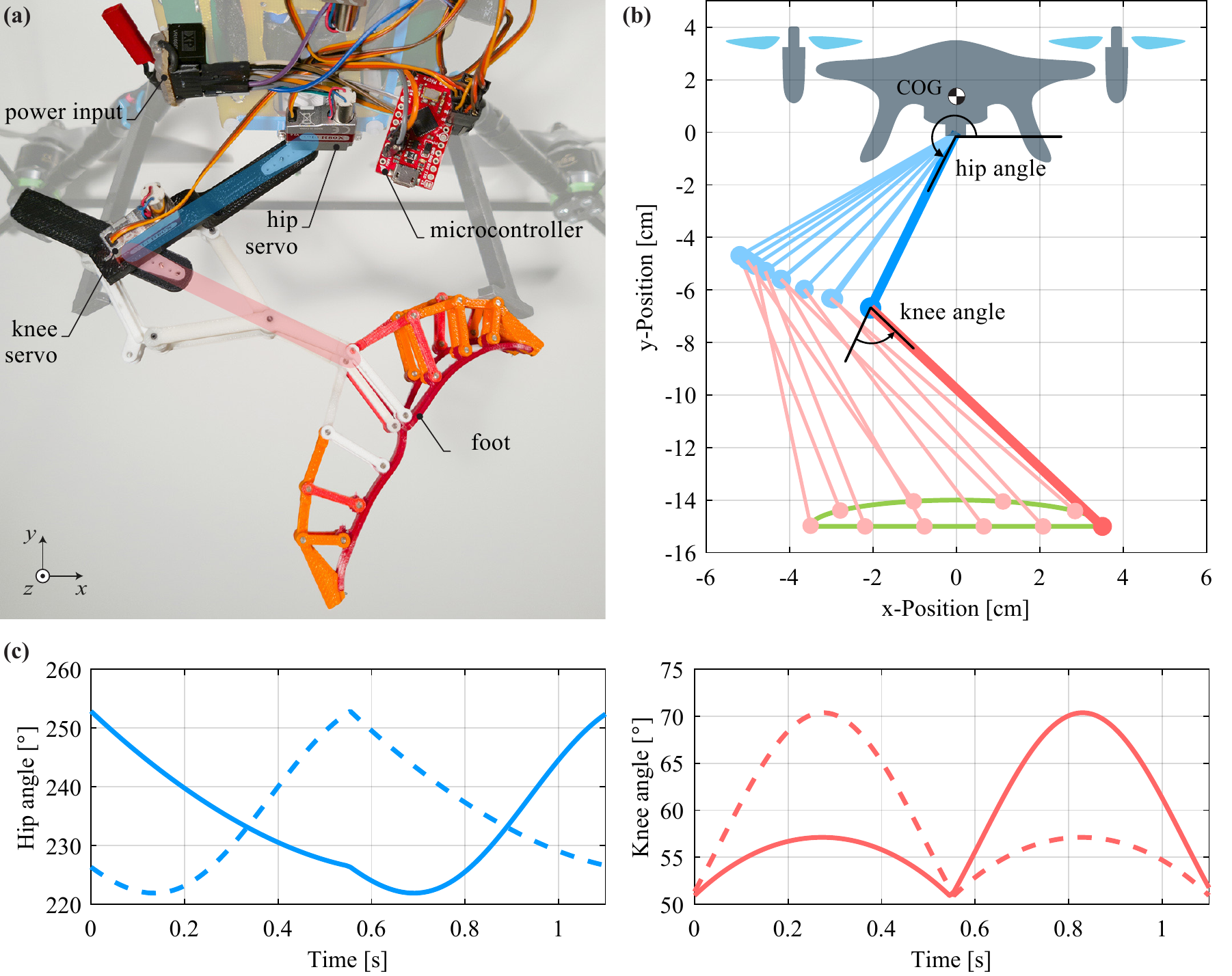}
\vspace{-0.55cm}
\caption{\textbf{(a)} Close-up picture of the leg and walking control system. \textbf{(b)} Leg and foot trajectories, illustrating the positions of the upper limb, lower limb, and foot for a walking cycle. The UAV is not drawn to scale. \textbf{(c)} Joint angles for the hip and knee joints. Solid and dashed lines represent left and right leg trajectories with a 50\% duty cycle phase difference.}
\vspace{-0.55cm}
\label{walking_trajectory}
\end{center}
\end{figure*}

\begin{figure*}[tbp]
\centering
\begin{center}
\includegraphics[width=1.0\textwidth]{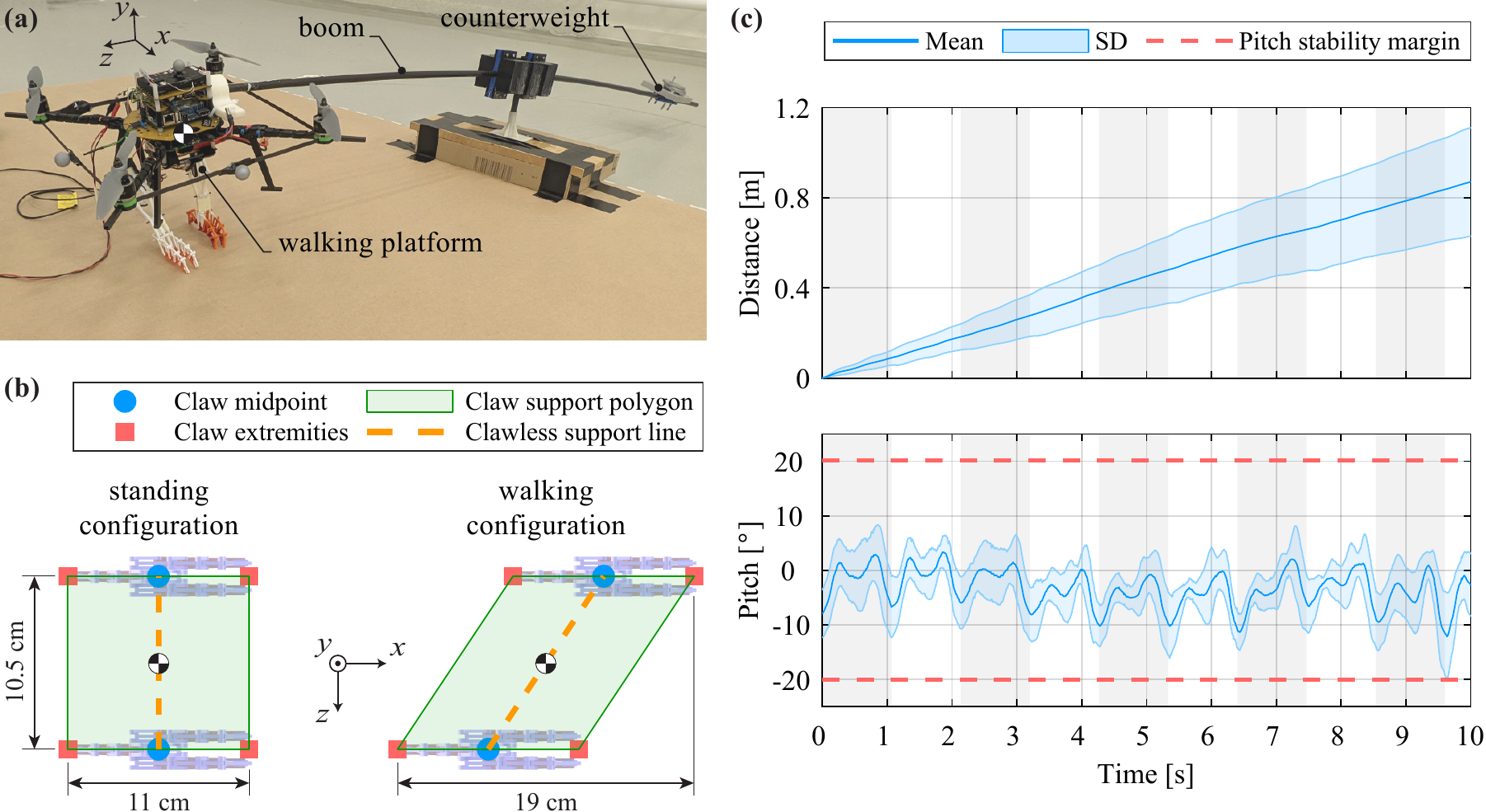}
\vspace{-0.4cm}
\caption{\textbf{(a)} Image of the walking experimental setup. \textbf{(b)} Support polygons with hyperextended claws, the left is when the two feet are at the maximum separation distance, and the right is when the legs are in a natural standing position. \textbf{(c)} Distance traveled and pitch angle data of walking experiments with the hyperextended feet. The grey-shaded regions indicate each gait cycle.}
\vspace{-0.5cm}
\label{walking_characterization}
\end{center}
\end{figure*}

A custom-built, manually controlled quadcopter UAV was equipped with two claws with talons, weighing \SI{700}{\gram} in total. To demonstrate perching in action, we conducted eight experiments with the UAV taking off from the ground, briefly flying around, and carefully landing on a \SI{40}{\milli\meter} diameter birch perch. The UAV successfully perched in six trials, only losing balance twice due to a high tilting angle resulting from a correspondingly high approach angle. To unperch, the UAV generates thrust from all four motors. As this happens, horizontal force that produces the squeezing decreases, and the claw releases the perch. When the thrust of the motors is greater than the weight of the UAV, it is free to move upward. Following perching, we successfully demonstrated unperching in four trials, demonstrating the claws' multi-functionality (see Supplementary Video). Figure \ref{perch_test} shows representative data from a successful perching trial. It also presents data on vertical distance (altitude) from the perch, the pitch angle (corresponding to $\theta$ in Fig. \ref{tilting_characterization}a-b), and snapshots of notable events during the perching maneuver. The experiments were carried out indoors, and the data were collected using an OptiTrack motion capture system.

We also investigated the limits of UAV tilting. To do this, the UAV was set on the birch perch and slowly pushed by hand until it fell off. After conducting the test 10 times, the UAV slipped at an average of 19.4 degrees with a standard deviation of 0.65 degrees. To further characterize the claw's versatility and compliance, we expanded our experiments to include irregular perches. This involved using a 3D-printed ABS circular perch and polygonal-shaped perches (triangle, square, pentagon, and star), all dimensioned to fit within a \SI{40}{\milli\meter} circle -- the same size as the tested birch perch.

Two key factors influence the performance of the claw on irregular perches: contact area and the engagement of talons. The claw's flexible toe pad conforms better to convex shapes, particularly those closer to the circular design (Fig. \ref{irregular perches}). By increasing the polygon sides, as we transition from a triangle to a pentagon and ultimately to a circle, the contact area increases, resulting in greater frictional surface and an overall improved tilting performance. However, talon engagement also plays a crucial role. In the case of the pentagon with a flat bottom, talons randomly get caught at the bottom side edges, resulting in an increased sustained angle with a large standard deviation. On concave-shaped perches like the star, talons have a more pronounced effect. On the regular star, the claw completely locked on the perch at around \SI{28}{\degree} as the talons fully engaged at the bottom. However, on the upside-down star perch, the talons failed to engage at all, resulting in poor performance due to minimal contact area.


\section{Legged Locomotion Characterization}

To demonstrate the stable walking ability of the proposed leg and claw system, we added actuation to the claws attached to the quadcopter used for the perching experiment (Fig. \ref{walking_trajectory}a). For each leg, two servo motors (KST X08), located directly at the joints, drive the hip and knee joints. The upper and lower limb lengths are 7 cm and 10 cm, respectively. The lower limb length, determined considering the torque limitation of the servo motors, cannot be shorter than 10 cm due to the length of the Hoberman linkage and the claw. The distance between the two legs is 10.5 cm. The walking gait is generated by a position control method following a half ellipse trajectory as shown in Fig. \ref{walking_trajectory} with a phase difference of 50\%. The flat trajectory is to provide a flat center of gravity (COG) forward movement when the foot is in contact with the ground, and the arc trajectory is to retract the leg without touching the ground. A Sparkfun Pro Micro board maps the half ellipse trajectory to the angular position values and commands the servo motors to follow the given angular position values.

The claw design can provide a stable gait for a bipedal system when the claw is in the hyperextended mode. The hyperextended feet provide extra contact points with the ground and enlarge the support polygon (Fig. \ref{walking_characterization}b). Fig. \ref{walking_characterization} shows the support polygon region formed by the extremities of the claws when both legs are on the ground. The Left and right diagrams present the support polygons when the legs are in a neutral position and when the legs are fully stretched forward and backward, respectively. The orange dashed lines indicate where the center of gravity would need to be located in the absence of the claws if the legs were point feet. With a 8 cm stride length, the support polygon changes from a rectangular stance area of \SI{99.75}{\centi\meter\squared} to a parallelogram of \SI{115.5}{\centi\meter\squared} during walking.

\begin{figure*}[ht]
\centering
\begin{center}
\includegraphics[width=1.0\textwidth]{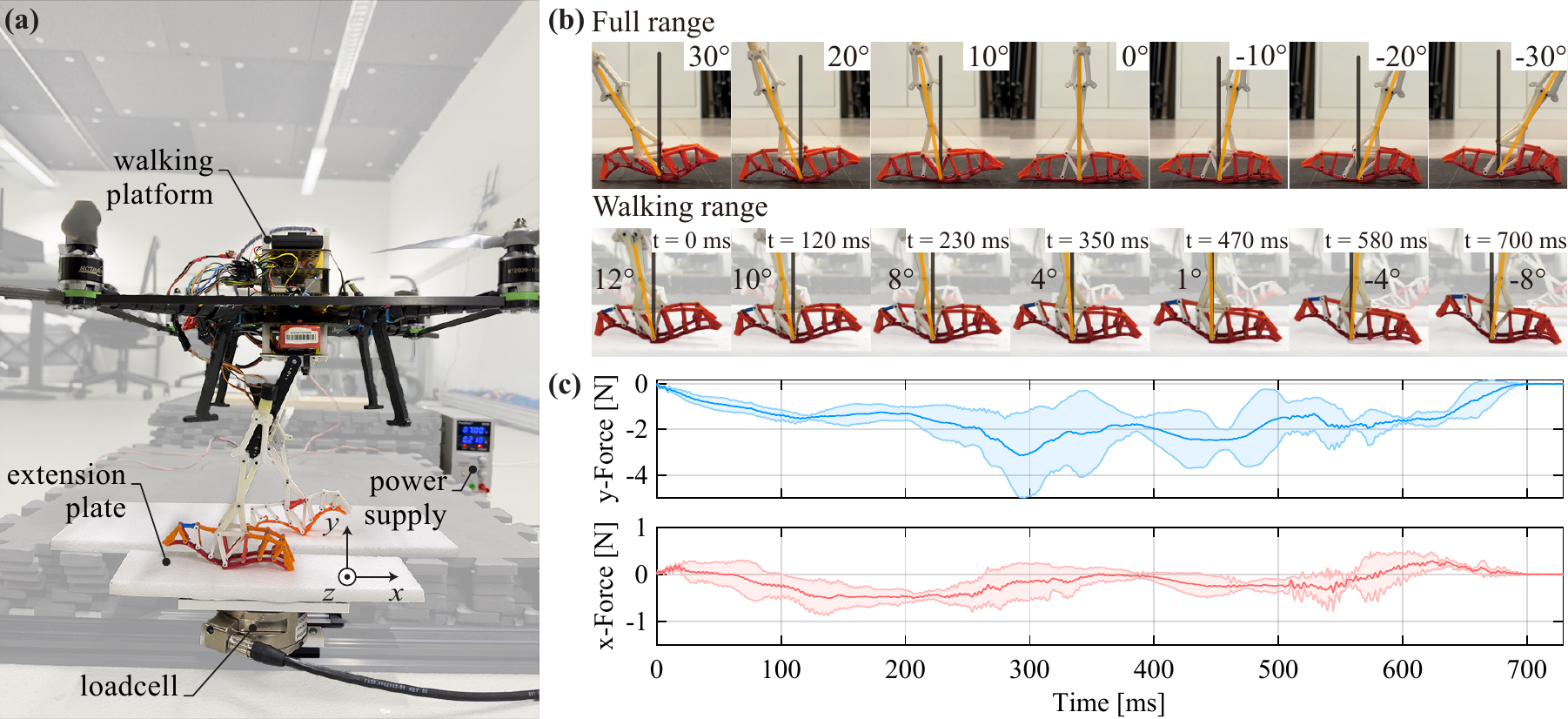}
\vspace{-0.4cm}
\caption{\textbf{(a)} The walking experimental setup with force sensor. \textbf{(b)} Claw deformation with different leg angles. \textbf{(c)} Single leg force data during walking.}
\vspace{-0.4cm}
\label{walking_deformation}
\end{center}
\end{figure*}

The robotic platform was connected to a boom support system and walked around a pivot axis in a circular trajectory (Fig. \ref{walking_characterization}). In the context of robotic locomotion, the use of a boom support is a common intermediate solution for evaluating and testing walking in legged robots \cite{pratt1998intuitive, hubicki2016atrias, shin2022elastic}. The boom system was installed for two reasons. The servo motors provide degrees of freedom only in the sagittal plane. Thus, the current setup could only provide stability in pitch angle. Another reason was to reduce the resultant weight with a counterweight due to the power limitation of the servo motors that drove the leg mechanism. The position and orientation of the robot were collected using an OptiTrack motion capture system to evaluate whether the claws' hyperextended configuration can provide stable terrestrial locomotion. Five walking trials were recorded at \SI{120}{\hertz}. On the other end of the boom, weights were installed to counterbalance the system, and the resultant effective weight of the system was \SI{200}{\gram}. The distance traveled data are presented in Fig. \ref{walking_characterization}c. The platform covered \SI{0.87}{\meter} in 10 seconds at \SI{1.1}{\hertz} of walking frequency with the hyperextended feet. The average power consumption during walking exceeded the energy expenditure for standing by more than threefold. Specifically, the energy consumed for one complete walking cycle was measured at \SI{2.34}{\joule}, whereas the energy expended during standing amounted to \SI{0.71}{\joule} over an identical period of time. It could not walk or even stand when the claws were closed due to the limited support polygon area. Figure \ref{walking_characterization}c also presents the pitch angle data during the walking, which shows a repeating pattern for each cycle. The pitch angle started around -8 degrees and had two local maximums of around 0 degrees for each step. This pattern was repeatedly shown for every walking cycle, and the average pitch stayed around -5 degrees during the 10 seconds of walking. The pitch angle range did not diverge from the start of the experiment but stayed between -12 degrees and 5 degrees. The stability margin is calculated based on the center of gravity position and the support polygon size. It indicates the maximum pitch angle beyond which the robot tips over. The maximum allowed pitching angle of the platform was $\pm$20.1 degrees, and the platform stayed within this range throughout the experiment. These results indicate that the hyperextended configuration provides stable walking in pitch because the average pitch angle does not change significantly in the 10 seconds of walking.

We also studied claw deformation during walking and the corresponding force profiles, as shown in Fig. \ref{walking_deformation}. We installed a load cell on the ground and attached an extension plate onto it to enable the walking platform to take at least two steps. The right leg walked on the extension plate, while the left leg walked on a separate surface of equal height. This arrangement ensured that the load cell recorded forces from a single leg only. For analysis, we focused on data from the second step to avoid potential impact noises from dropping the platform onto the plate during the first step. The duration of ground contact was \SI{700}{\milli\second}. The gradual vertical force transitions during the initial and final \SI{100}{\milli\second} indicate the compliance of the claw. The horizontal force profile reveals that the forward force was primarily generated within the first \SI{300}{\milli\second}. Positive horizontal forces were observed at the beginning and end, possibly due to unintended contact with the ground during the foot's swing phase. The theoretical duration of the stance phase of the walking platform is \SI{450}{\milli\second}, which aligns with the duration of the measured negative horizontal force.

During walking, the leg angle ranged from -8 degrees to 12 degrees with the stride length of \SI{8}{\centi\meter}. In a separate experiment shown in the top row of Fig. \ref{walking_deformation}b, the maximum leg angle range and corresponding claw deformation were evaluated. The maximum range observed was from -25 degrees to 25 degrees. As the leg tilted away further from the vertical position, the curvature between the middle point and the extremity increased. When the claw reached the maximum angle, it stopped deforming, and the middle point of the sole was lifted off the ground, as depicted in the rightmost snapshot.


\section{Conclusions}

The claws presented in this manuscript demonstrate a new solution by which a UAV can perch and walk, using a reconfigurable mechanism. The lightweight construction of the Hoberman linkage legs and \finray claws are capable of holding up to a \SI{700}{\gram} UAV. We have presented important sizing relationships and static models for calculating expected performance.

Although the claws developed here are completely passive, actuators could be used to further increase their squeezing force allowing them to be scaled up for use on heavier UAVs. These actuators would mimic the muscles that birds use to grasp perches. Future iterations could also include actuators, such as a motor and tendon, to automatically pass through singularity and, consequently, switch leg modes between walking and perching. 

These claws widen the range of possibilities of multimodal UAVs by enabling walking and perching. In particular, the use of these claws in semi- or fully-erect walking robots gives them advantages over sprawling or wheeled alternatives in navigating cluttered environments. This is because their longer legs are able to stride over larger obstacles. The perching capability allows them to remain in position for long periods of time to recharge batteries, conduct observations, or minimize noise and power consumption. This will lead to future robots with a larger range of capabilities, making them more versatile tools for search and rescue operations.

\section{Acknowledgements}
The authors are grateful for the engineering help provided by Olexandr Gudozhnik, the useful feedback from Florian Achermann, and the piloting skills of Przemyslaw Kornatowski, Victor C. Rochel, Nathan S. Müller, and Simon Jeger.


\bibliography{IEEEabrv,EPFLBib}

\begin{IEEEbiography}[{\includegraphics[width=1in,height=1.25in]{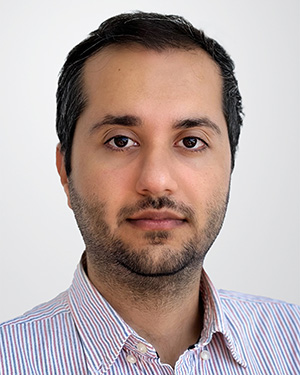}}]{Mohammad Askari}
received the B.S. (2015) and M.S. (2018) degrees, both with the highest distinction, in mechanical engineering from Middle East Technical University and Bilkent University in Ankara, Turkey, respectively. From 2015 to 2019, he worked on origami-inspired miniature legged robots at the Bilkent Miniature Robotics Laboratory. Since 2020, he has been with the Laboratory of Intelligent Systems, Swiss Federal Institute of Technology Lausanne (EPFL), Lausanne, Switzerland, focusing on mechanical design and control of aerial robots.
\vspace{-0.1cm}
\end{IEEEbiography}%

\begin{IEEEbiography}[{\includegraphics[width=1in,height=1.25in]{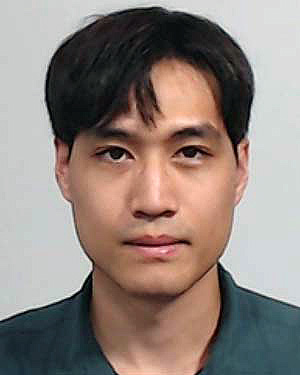}}]{Won Dong Shin}
received the B.S. and M.S. degrees in mechanical engineering from the University of Illinois at Urbana-Champaign, Champaign, IL, USA, in 2016 and 2018, respectively. Since 2019, he has been working toward the Ph.D. degree in robotics with the Laboratory of Intelligent Systems, \'{E}cole Polytechnique F\'{e}d\'{e}rale de Lausanne (EPFL), Lausanne, Switzerland.
\end{IEEEbiography}%

\begin{IEEEbiography}[{\includegraphics[width=1in,height=1.25in]{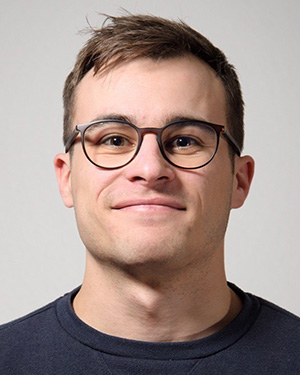}}]{Damian Lenherr}
received the B.S. and M.S. degrees in mechanical engineering from the Swiss Federal Institute of Technology in Zurich (ETHZ), Zurich, Switzerland, in 2019 and 2021, respectively. As part of a student exchange program during the spring semester of 2021, he conducted his master's thesis research at the Laboratory of Intelligent Systems at the Swiss Federal Institute of Technology Lausanne (EPFL), Lausanne, Switzerland.
\end{IEEEbiography}%

\begin{IEEEbiography}[{\includegraphics[width=1in,height=1.25in]{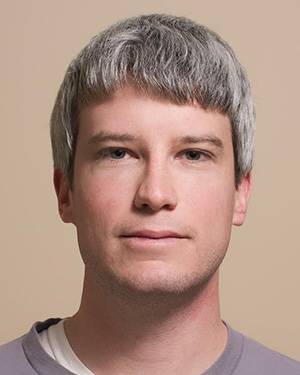}}]{William Stewart (Member, IEEE)}
received the M.S. and Ph.D. degrees in aerospace engineering from North Carolina (NC) State University, Raleigh, NC, USA, in 2014 and 2018, respectively. From 2010 to 2014, he was a leader of the NC State University Aerial Robotics Team. Between 2018 and 2022, he worked as a Postdoctoral Researcher at the Laboratory of Intelligent Systems, Swiss Federal Institute of Technology Lausanne (EPFL), Lausanne, Switzerland. He is currently an Assistant Professor at Stony Brook University. His research interests include aerodynamics, mechanics, and vehicle design. Dr. Stewart was two-time AUVSI student UAS competition champion (2010 and 2014).
\end{IEEEbiography}%

\begin{IEEEbiography}[{\includegraphics[width=1in,height=1.25in]{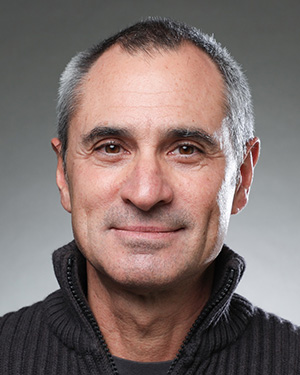}}]{Dario Floreano (Fellow, IEEE)}
received the M.A. degree in Visual Psychophysics from Univ. Trieste, Italy, in 1988, the M.S. degree in Neural Computation from Univ. Stirling, UK, in 1992, and the Ph.D. degree in Evolutionary Robotics from Univ. Trieste, Italy, in 1995. He is a full Professor and Director of the Laboratory of Intelligent Systems at EPFL, Switzerland. He was the Founding Director of the Swiss National Center of Competence in Robotics from 2010 to 2022. He held visiting fellowships with Sony Computer Science Laboratory, Tokyo, Japan; Caltech/JPL, Pasadena, CA, USA; and Harvard University, Boston, MA, USA. His research interests include biologically inspired robotics and artificial intelligence. He has made pioneering contributions to the fields of evolutionary robotics, aerial robotics, and soft robotics. He is on the advisory board of several institutes and international organizations in robotics and artificial intelligence and has served on the editorial board of several scientific journals.
\end{IEEEbiography}%


\clearpage
\noindent\hfill\textbf{\large{Supplementary Text}}\hfill
\renewcommand{\thefigure}{S\arabic{figure}}
\setcounter{figure}{0}

\begin{appendices}

\section*{Claw Linkage Design and Fabrication}
\label{app:linkage}

We determine the sizing of the finray geometry in the perched configuration by allowing the digits to overlap by \SI{90}{\degree} on the minimum perch diameter (see Fig. \ref*{claw}a in the main text). Positioning the toe pad hubs at angles of \SI{60}{\degree}, \SI{115}{\degree}, \SI{165}{\degree}, and \SI{225}{\degree} relative to the vertical axis around the perch allows for this arrangement. The ribs are then connected to the outer-links, which are positioned radially above the subsequent joints on the toe pad. The length of the ribs has to be long enough, so that they do not intercept with the sole or the toe pad hubs. To ensure an adequate force transfer through the outer wall, it is necessary to minimize the relative angle of the subsequent outer-links.

\begin{figure}[htb]
\centering
\includegraphics[draft=false,width=\columnwidth]{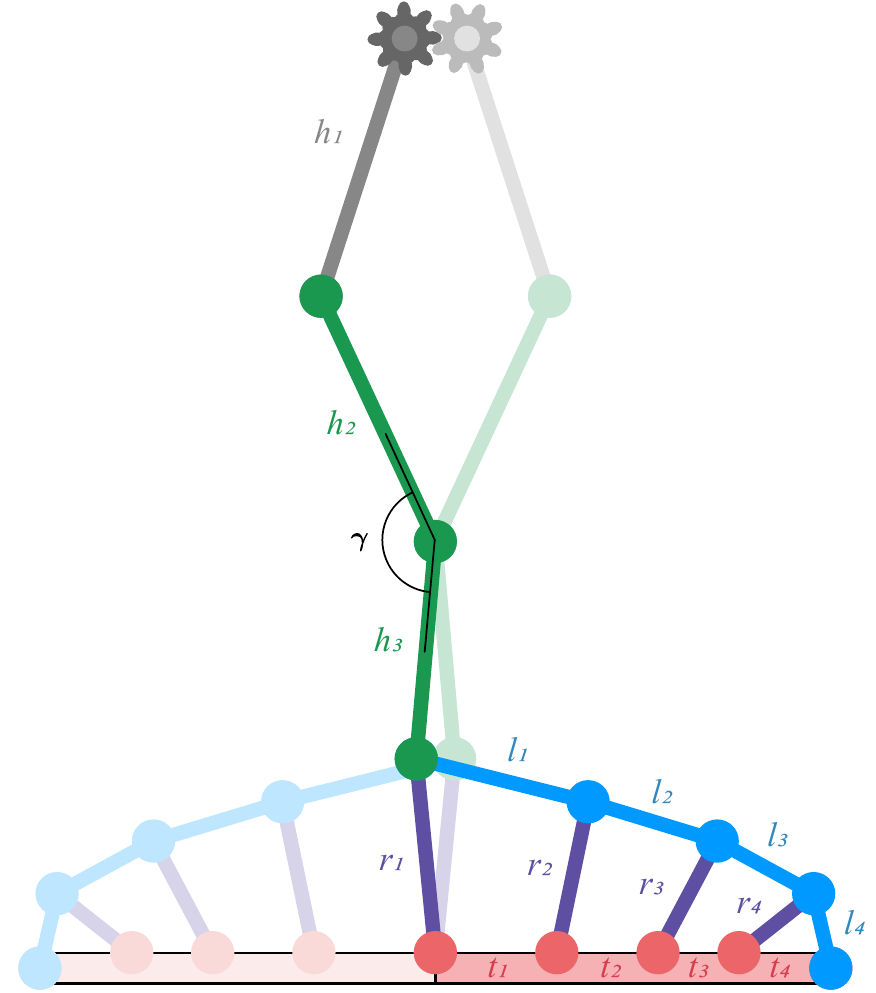}
\caption{Schematic diagram of the claw in flat configuration. The lowercase letters represent the link lengths for Hoberman ($h$), toe pad ($t$), ribs ($r$), and outer-links ($l$). $\gamma$ is the fixed Hoberman angle from Fig. \ref*{claw}c of the main text. Due to symmetry, only half of the claw system is labeled.}
\label{fig:flat claw}
\end{figure}

Once the link lengths of the \finray claw are determined in the perched configuration, the foot is designed for fabrication in the flat configuration (Fig. \ref{fig:flat claw}). This allows the toe pad to be 3D-printed as a single flat piece containing all the hubs. We print it out of a flexible TPU material (NinjaFlex 85A – with an elastic modulus of $E=\SI{12}{\mega\pascal}$) using a Prusa MK3s printer. The sole of each digit is tapered, narrower at the tip than at the base, to allow sufficient space for the overlapping of toe tips. All other links are 3D-printed using ABS ($E=\SI{1.6}{\giga\pascal}$) on a Stratasys Dimension Elite printer. Table \ref{tab:linkage} enlists all the link lengths for the complete claw system, based on the chosen minimum perch diameter of \SI{30}{\milli\meter}.

\begin{table}
\newcommand{\ANG}[1]{\SI{#1}{\degree}}
\centering
\caption{Dimensions of the full claw system (\SI{}{\milli\meter})}
\begin{tabularx}{\columnwidth}{@{}lrXlrXlrXlr@{}}
\toprule
\multicolumn{2}{c}{Hoberman} && \multicolumn{2}{c}{outer-links} && \multicolumn{2}{c}{ribs} && \multicolumn{2}{c}{toe pad} \\
\cmidrule{1-2} \cmidrule{4-5} \cmidrule{7-8} \cmidrule{10-11}
$h_1$    & 35.00     && $l_1$ & 22.91 && $r_1$ & 25.19 && $t_1$ & 15.71 \\
$h_2$    & 35.00     && $l_2$ & 17.45 && $r_2$ & 19.93 && $t_2$ & 13.09 \\
$h_3$    & 28.22     && $l_3$ & 14.21 && $r_3$ & 16.34 && $t_3$ & 10.47 \\
$\gamma$ & \ANG{150} && $l_4$ &  9.89 && $r_4$ & 12.31 && $t_4$ & 11.88 \\
\bottomrule
\end{tabularx}
\label{tab:linkage}
\end{table}

Note that for proper operation of the Hoberman linkage, the link length $h_3$ must satisfy $h_3 > r_1$. This ensures that there is no contact between the Hoberman and toe pad joints when the mechanism folds, preventing any hindrance to force transmission through the outer-links. In order to accommodate symmetric movement of all the digits, the left and right Hoberman linkages are connected at the top to each other via a set of 3D-printed (ABS) gears of \SI{4}{\milli\meter} radius. The toe pad's sole is made \SI{4}{\milli\meter} thick.  Each Hoberman, rib, and outer-link has a cross section of \SI{2}{\milli\meter} by \SI{3}{\milli\meter} and is equipped with a plain bearing at each end, designed to accommodate steel pins or carbon tubes with a diameter of \SI{2}{\milli\meter}. To create a clearance fit, the holes are drilled out with a \SI{2.1}{\milli\meter} drill bit.

\section*{Perching Model}
\label{app:model}

\subsection{Model Assumptions}

In modeling the claw for perching (Fig. \ref{fig:fbd perched}), we make the following assumptions:

\begin{figure}[htb]
\centering
\includegraphics[draft=false,width=\columnwidth]{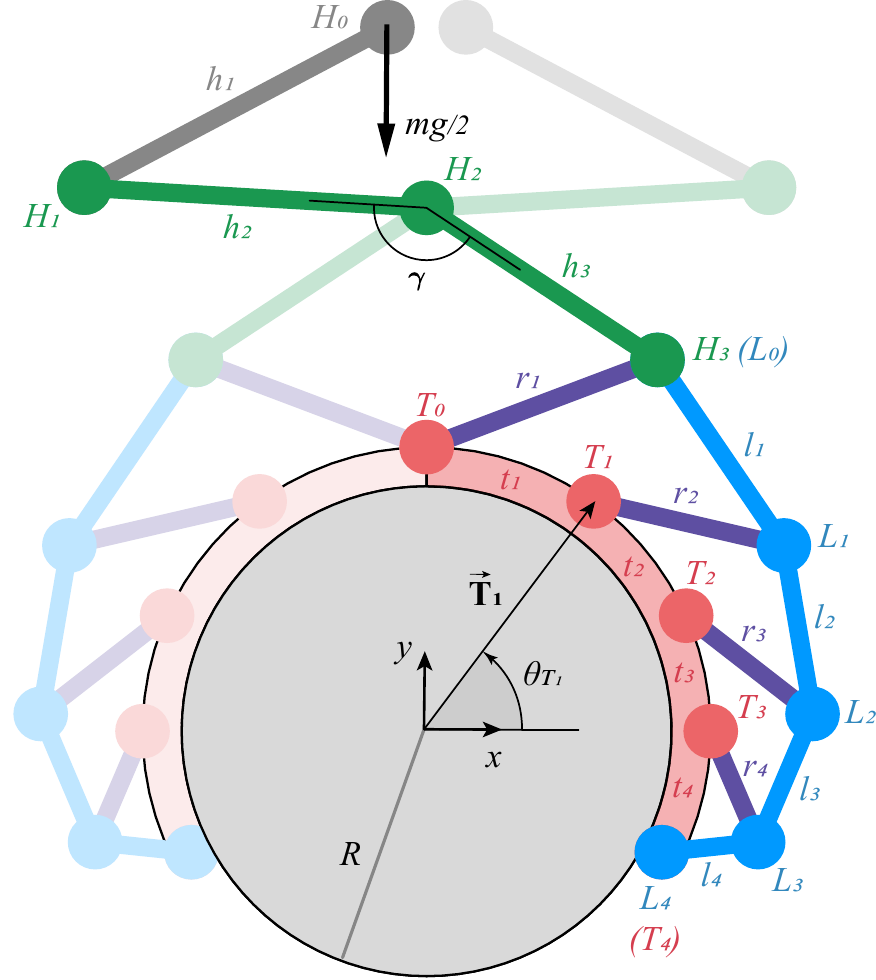}
\caption{Schematic diagram of the perched claw on a \SI{50}{\milli\meter} perch. Capital letters denote the joint names, and lowercase letters the corresponding link lengths. An example of a vector notation representing the position vector to a joint is shown ($\protect\vecsub{T}{1}$), with its angle denoted by $\theta$ preceding the joint name.}
\label{fig:fbd perched}
\end{figure}

\begin{enumerate}

\item Under any payload, the claw's soft sole (red toe pad) conforms to the shape of the perch of any size and is in perfect contact with it.

\item Due to the claw's planar symmetry, we only consider modeling the linkage on one side. Furthermore, the payload is assumed to be split equally between the two sides.

\item The claw in perched configuration is treated as a fully rigid system with negligible weight. This means that all the comprising links, including the soft toe pad, are assumed to be rigid and weightless.

\item The model does not include the talons and only predicts the talon-less claw type.

\end{enumerate}

\subsection{Kinematic Analysis}

For a given perch radius of $R$ and sole thickness of $S$, we first solve for all the joint locations in perched configuration. The position vector of the toe pad joints are found by:

\begin{equation*}
\vecsub{T}{i} = (R + S) \begin{bmatrix} \cos{\theta_{T_i}} & \sin{\theta_{T_i}} & 0 \end{bmatrix}^\top ,
\end{equation*}
\begin{equation*}
\mathrm{where \ } \theta_{T_i} = \frac{\pi}{2} - \frac{1}{R} \sum_{j=0}^i t_j \ .
\end{equation*}

In the above equations, $i$ is the subscript number of the toe pad joint ($i=0$ to $4$), and $\theta_{T_i}$ represents the angle of the position vector from horizontal. If we calculate and denote the angle of $(\vecsub{T}{4} - \vecsub{T}{3})$ vector by $\theta_{t_4}$, the position of joint $T_3$ can be found using the cosine law for triangle $T_3 T_4 L_3$ as follows:

\begin{equation*}
\vecsub{L}{3} = \vecsub{T}{3} + r_4 \begin{bmatrix} \cos{(\theta_{t_4}+\alpha)} & \sin{(\theta_{t_4}+\alpha)} & 0 \end{bmatrix}^\top ,
\end{equation*}
\begin{equation*}
\mathrm{where \ } \alpha = \arccos{ \frac{t_4^2 + r_4^2 - l_4^2}{2 t_4 r_4} } \ .
\end{equation*}

The rest of the \finray linkage has three interdependent four-bars, the kinematic solution of which can yield the position of the remaining outer-link joints ($L_2$, $L_1$, and $L_0$). These four-bar linkages are $T_2 T_3 L_3 L_2$, $T_1 T_2 L_2 L_1$, and $T_0 T_1 L_1 L_0$. We skip here detailed derivation of these joint locations for brevity and present only the method used to solve the four-bar kinematics in the following section.

Having determined all the joint locations of the outer-links, we can solve for the Hoberman joints as follows:

\begin{align*}
\vecsub{H}{2} &= \begin{bmatrix} 0 & L_{3y} + \sqrt{h_3^2 - L_{0x}^2} & 0 \end{bmatrix}^\top , \\
\vecsub{H}{1} &= \vecsub{H}{2} + h_2 \begin{bmatrix} \cos{(\theta_{h_3}-\gamma)} & \sin{(\theta_{h_3}-\gamma)} & 0 \end{bmatrix}^\top , \\
\vecsub{H}{0} &= \begin{bmatrix} 0 & H_{1y} + \sqrt(h_1^2 - (H_{0x}-H_{1x})^2) & 0 \end{bmatrix}^\top ,
\end{align*}

\noindent where $\theta_{h_3}$ is the angle of $(\vecsub{H}{3}-\vecsub{H}{2})$ vector relative to positive $x$ axis and $H_{0x}$ is the fixed horizontal distance corresponding to the radius of the gear at the $H_0$ joint.

\subsection{Four-bar Position Kinematics}

\begin{figure}[htb]
\centering
\includegraphics[draft=false,width=\columnwidth]{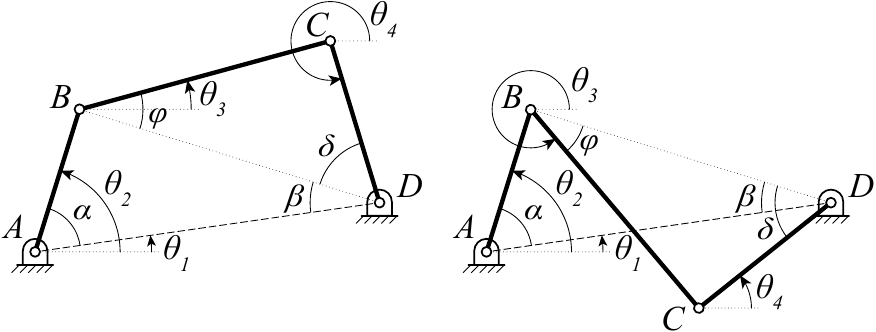}
\caption{The possible solutions for a generic four-bar mechanism with fixed pivots at $A$ and $D$: open configuration (left) and crossed configuration (right).}
\label{fig:generic four-bar}
\end{figure}

The position analysis of four-bar mechanisms generally consists of nonlinear equations, which can be computationally challenging or slow to solve. This nonlinearity arises due to the existence of multiple solutions (Fig. \ref{fig:generic four-bar}). We propose here a unique position analysis method, applicable to any four-bar mechanism, to avoid solving any nonlinearity. The solution presented here is based on the method proposed by \cite{mata2016mechanisms}.

For any given crank input ($\theta_2$) to a generic four-bar mechanism shown in Fig. \ref{fig:generic four-bar}, with known link lengths and fixed pivot positions of $A$ and $D$, we can solve for the unknown coupler and rocker angles, $\theta_3$ and $\theta_4$, respectively, as follows:

\begin{align*}
\alpha &= \theta_2 - \theta_1 \ , \\
\overline{BD}     &= \sqrt{\overline{AB}^2 + \overline{AD}^2 - 2 \, \overline{AB} \, \overline{AD} \, \cos \alpha} \ , \\
\beta  &= \mathrm{atan2} \big( \sin \beta, \cos \beta \big) \ ,
\end{align*}
where,
\begin{equation*}
\sin \beta = \frac{\overline{AB}}{\overline{BD}} \sin \alpha \quad \mathrm{and} \quad \cos \beta= \frac{\overline{AD}^2 + \overline{BD}^2 - \overline{AB}^2}{2 \, \overline{AD} \, \overline{BD}} \ .
\end{equation*}
Similarly,
\begin{align*}
\phi   &= \arccos \Big( \frac{\overline{BC}^2 + \overline{BD}^2 - \overline{CD}^2}{2 \, \overline{BC} \, \overline{BD}} \Big) \ , \\
\delta  &= \mathrm{atan2} \big( \sin \delta, \cos \delta \big) \ ,
\end{align*}
where,
\begin{equation*}
\sin \delta = \frac{\overline{BC}}{\overline{CD}} \sin \phi \quad \mathrm{and} \quad \cos \delta= \frac{\overline{BD}^2 + \overline{CD}^2 - \overline{BC}^2}{2 \, \overline{BD} \, \overline{CD}} \ .
\end{equation*}

Then the solutions to the two possible configurations of a four-bar are given by:
\begin{equation*}
\text{\makebox[\widthof{Crossed configuration: }][r]{Open configuration: }}
\begin{cases}
\theta_3 = \theta_1 - \beta + \phi   \ , \\
\theta_4 = \theta_1 - \beta - \delta \ ,
\end{cases}
\end{equation*}
\begin{equation*}
\text{Crossed configuration: }
\begin{cases}
\theta_3 = \theta_1 - \beta - \phi   \ , \\
\theta_4 = \theta_1 - \beta + \delta \ .
\end{cases}
\end{equation*}

Note that $\mathrm{atan2}(y,x)$ in the above equations is the four-quadrant inverse tangent function of the two variables $x$ and $y$, i.e, $\arctan(x/y)$.

\subsection{Static Force Equilibrium}

Following the kinematic analysis of the claw in perched configuration, a static force balance is carried out. Doing so can resolve the internal forces at each joint based on the input payload. The horizontal components of the resolved forces at the toe pad joints ($T_1$ through $T_4$ in Fig. \ref{fig:fbd perched}) can then be compared to the split perch test results for validation.

A total of $40$ unknown forces are considered when formulating the linear system of equations. We place two reaction forces at joints connected by two links or less and six reaction forces for those connected by three links ($L_0$ through $L_3$), i.e., two forces per each of the three links. Hence, a minimum of $40$ equations are needed to resolve these forces. Note that enforcing a two-force member assumption can greatly reduce the number of unknowns. Except for the Hoberman link, which has three joints, all the links are under compression or tension. Hence the resultant reaction forces at their joints are colinear, i.e., being a two-force member. However, we do not enforce this and instead use it as a tool to verify whether the more general approach satisfies the two-force member assumption once solved. $33$ equations come from force and moment balance about the $11$ rigid links in the system (including the whole toe pad as one link). The remaining ($8$ equations) come from a force equilibrium constraint at the outer wall joints connected by three links, resulting in a total of $41$ equations.

For brevity, we do not derive the entire system of equations here and present only a few examples for the Hoberman links and outer-link joint $L_0$. If we define $\textrm{moment}(\vec{r},\vec{F})$ as the z-component of the cross product of $\vec{r} \times \vec{F}$, then we have from force equilibrium of $h_1$ link:

\begin{align*}
\textrm{eqn1: } & +F_{H_{0x}} + F_{H_{1x}} = 0 \ , \\
\textrm{eqn2: } & +F_{H_{0y}} + F_{H_{1y}} -mg/2 = 0 \ , \\
\textrm{eqn3: } & \textrm{moment}(\vec{r},\begin{bmatrix} +F_{H_{1x}} & 0 & 0 \end{bmatrix}^\top) \ + \\
                & \textrm{moment}(\vec{r},\begin{bmatrix} 0 & +F_{H_{1y}} & 0 \end{bmatrix}^\top) = 0 \ , \\
\textrm{where } & \vec{r} = \vecsub{H}{1} - \vecsub{H}{0} \ .
\end{align*}

Similarly, the force equilibrium of $h_2 h_3$ link yields:

\begin{align*}
\textrm{eqn4: } & -F_{H_{1x}} + F_{H_{2x}} + F_{H_{3x}} = 0 \ , \\
\textrm{eqn5: } & -F_{H_{1y}} + F_{H_{2y}} + F_{H_{3y}} = 0 \ , \\
\textrm{eqn6: } & \textrm{moment}(\vecsub{r}{1},\begin{bmatrix} -F_{H_{1x}} & 0 & 0 \end{bmatrix}^\top) \ + \\
                & \textrm{moment}(\vecsub{r}{1},\begin{bmatrix} 0 & -F_{H_{1y}} & 0 \end{bmatrix}^\top) \ + \\
                & \textrm{moment}(\vecsub{r}{2},\begin{bmatrix} +F_{H_{3x}} & 0 & 0 \end{bmatrix}^\top) \ + \\
                & \textrm{moment}(\vecsub{r}{2},\begin{bmatrix} 0 & +F_{H_{3y}} & 0 \end{bmatrix}^\top) = 0 \ , \\
\textrm{where } & \vecsub{r}{1} = \vecsub{H}{1} - \vecsub{H}{2} \textrm{ and } \vecsub{r}{2} = \vecsub{H}{3} - \vecsub{H}{2} \ .
\end{align*}

As an example of the force equilibrium constraint at outer wall joints with three connected links, we have at joint $L_0$:

\begin{align*}
\textrm{eqn7: } & +F_{H_{3x}} + F_{L_{0x}}^{r_1} + F_{L_{0x}}^{l_0} = 0 \ , \\
\textrm{eqn8: } & +F_{H_{3y}} + F_{L_{0y}}^{r_1} + F_{L_{0y}}^{l_0} = 0 \ .
\end{align*}

The remaining static equilibrium equations can be derived with a similar approach. If we denote the vector of $40$ unknown forces by $\vec{x}$, then we can formulate all equations as a linear system in the form: $\mat{A}\vec{x} = \vec{b}$. Note that $\mat{A}$ is a $41 \times 40$ matrix, and $\vec{b}$ a $41 \times 1$ vector of external forces (payload only in this case). Hence, the forces at the joints are found by $\vec{x} = \mat{A}^{\!-1} \vec{b}$, where $\mat{A}^{\!-1}$ is the Moore-Penrose pseudoinverse of $\mat{A}$.

\subsection{Squeezing Force Estimation}

The claw can only press on the perch with the forces exerted by its toe pad joints. However, not all joints contribute to a squeezing force. To compare results with the split perch experiments, we estimate the total exerted squeezing force on one half of the perch by the toe pad using:

\begin{align*}
\def\arraystretch{2}
F_{sq} &= \sum_{i=1}^n \widehat{F}_{T_{ix}} \ , \\
\text{where } \ \widehat{F}_{T_{ix}} &= \left\{\def\arraystretch{1.2}\begin{array}{lr}
		\phantom{|F} 0	            & F_{T_{ix}}>0 \\
		\left| {F}_{T_{ix}} \right|	& \phantom{-}F_{T_{ix}}\leq0 \end{array}\right. .
\end{align*}

From the actual split perch experiments, we observed that the number of joints contributing to a squeezing force varies with the perch diameter. Therefore, a value of $n=2$, $3$, and $4$ were used for the \SI{30}{\milli\meter}, \SI{40}{\milli\meter}, \SI{50}{\milli\meter} perch, respectively.

\subsection{Maximum Tilting Moment and Sustainable Angle}

We estimate the maximum sustainable angle using an inverted pendulum model with friction. When perching at an angle, the weight of the UAV ($mg$ in Fig. \ref{tilting_characterization}b) is off-centered and creates a moment ($M_w$), which tends to cause the claw to slip. At the maximum sustainable angle, this moment must be balanced with a moment from the friction between the toe pad of the claw and the perch surface ($M_f$). We take an iterative approach to find the maximum tilting angle. Assuming an initial angle of $\theta_a$, the moment due to weight $M_w$ can be calculated using the following relation:

\begin{equation}\label{eq:massloads}
    M_w = (R+L) mg \sin \theta_a \ ,
\end{equation}

\noindent where $R$ is the radius of the perch and $L$ is the lever arm length from the center of gravity to the perch surface. To find the resisting moment due to friction, we need to find the forces at each toe pad joint of the tilted claw. The tilting causes an asymmetric behavior between the right and left digits. Therefore, the kinematic analysis and the static force equilibrium steps need to be repeated separately for the left and right digits with the assumed inclination angle of $\theta_a$. This gives the forces in the $xy$ Cartesian coordinates, which must be resolved into radial and tangential components. The radial (normal) components create friction between the toe pad and the perch. If we resolve and denote the radial components of the toe pad joints by $F_{T_{ir}}$, the total friction force and the opposing moment to weight can be estimated by:

\begin{align*}
\def\arraystretch{2}
M_f &= R \mu F_r \ , \\
F_r &= \sum_{digit}^{l,r} \sum_{i=0}^4 \widehat{F}_{T_{ir}} \ , \\
\text{where } \ \widehat{F}_{T_{ir}} &= \left\{\def\arraystretch{1.2}\begin{array}{lr}
		F_{T_{ir}}            & F_{T_{ir}}\geq0 \\
		\phantom{F} 0	  & \phantom{-}F_{T_{ir}}<0 \end{array}\right. .
\end{align*}

In the above equations, $i$ is the subscript number of the toe pad joint ($i=0$ to $4$), $\mu_s$ is the static coefficient of friction between the toe pad and perch, and $l$ and $r$ denote the left and right digits, respectively. The condition that yields the maximum sustainable angle is $M_w = M_f$. If there is a mismatch between the two opposing moments, the assumed tilting angle ($\theta_a$) must be updated, and the steps are repeated to find the solution iteratively.

\end{appendices}

\end{document}